\documentclass[11pt]{article} 

\usepackage{graphicx}
\usepackage{booktabs}
\usepackage{algorithm}
\usepackage{algpseudocode}
\usepackage[T1]{fontenc}
\usepackage{lmodern}
\usepackage[flushleft]{threeparttable}
\usepackage[hidelinks]{hyperref} %
\hypersetup{hidelinks}
\usepackage{multirow}
\usepackage{longtable}
\usepackage[margin=3cm]{geometry} %
\usepackage{amsmath}
\usepackage{authblk}
\usepackage{caption}
\usepackage{cite}

\title{The effects of Hessian eigenvalue spectral density type on the applicability of
Hessian analysis to generalization capability assessment of neural
networks}
\author[1,*]{Nikita Gabdullin}
\affil[1]{Joint Stock ``Research and production company ``Kryptonite'' \authorcr
E-mail: n.gabdullin@kryptonite.ru}
\affil[*]{corresponding author}
\date{}
\begin{document}

    \captionsetup[table]{labelformat={default},labelsep=period,name={Table}}

    \maketitle

    \begin{abstract}
        Hessians of neural network (NN) contain essential
        information about the curvature of NN loss landscapes which can be used
        to estimate NN generalization capabilities. We have previously proposed
        generalization criteria that rely on the observation that Hessian
        eigenvalue spectral density (HESD) behaves similarly for a wide class of
        NNs. This paper further studies their applicability by investigating
        factors that can result in different types of HESD. We conduct a wide
        range of experiments showing that HESD mainly has positive eigenvalues
        (MP-HESD) for NN training and fine-tuning with various optimizers on
        different datasets with different preprocessing and augmentation
        procedures. We also show that mainly negative HESD (MN-HESD) is a consequence
        of external gradient manipulation, indicating that the previously
        proposed Hessian analysis methodology cannot be applied in such cases.
        We also propose criteria and corresponding conditions to determine HESD
        type and estimate NN generalization potential. These HESD types and
        previously proposed generalization criteria are combined into a
        unified HESD analysis methodology. Finally, we discuss how HESD changes
        during training, and show the occurrence of quasi-singular (QS) HESD and
        its influence on the proposed methodology and on the conventional assumptions 
        about the relation between Hessian eigenvalues and NN loss landscape curvature.
    \end{abstract}

    \emph{Keywords}: Neural networks, Hessian analysis, loss landscape, generalization. 
    
    \section{Introduction}\label{introduction}

Generalization capabilities of neural networks (NNs) is an extremely
important topic in modern Machine Learning (ML). It is commonly assumed
that high generalization can be obtained by training NNs on large
(possibly augmented) datasets aimed at accurate representation of
significant data variability present ``in the wild''. However, since
there are no guarantees that models will generalize well, such approaches
require to train numerous models with various hyperparameters to achieve
satisfactory results, which makes them extremely computationally
expensive. Hence, analysis methods that could provide some insight into
generalization capabilities of NNs are very important. Such methods
include analyzing loss landscapes~\cite{LLO,LLe} and Hessian~\cite{4H,PyH} of NNs.

We previously discussed how analyzing changes in Hessian eigenvalue
spectral density (HESD) can be used to assess generalization
capabilities of NN classifiers~\cite{prev}. It was also shown that HESDs
of NNs mostly exhibit typical behavior during training, with negative
eigenvalue section gradually reducing in size while large outliers occur
in the positive section. We refer to such HESDs as mainly positive (MP-HESD)
in this paper. Furthermore, the growth in negative section of MP-HESD
can indicate poor generalization when a trained NN model is tested on a
different dataset. However, there are cases when NN HESD can mainly have
negative eigenvalues (MN-HESD), so the proposed criteria predict poor
generalization, but in reality, such NNs may still generalize
very well. In this paper we discuss the origins of MN-HESD and study
whether Hessian analysis methodology~\cite{prev} can be applied to
analyze such cases.

\begin{figure} 
    \centering
    \includegraphics[scale=0.38]{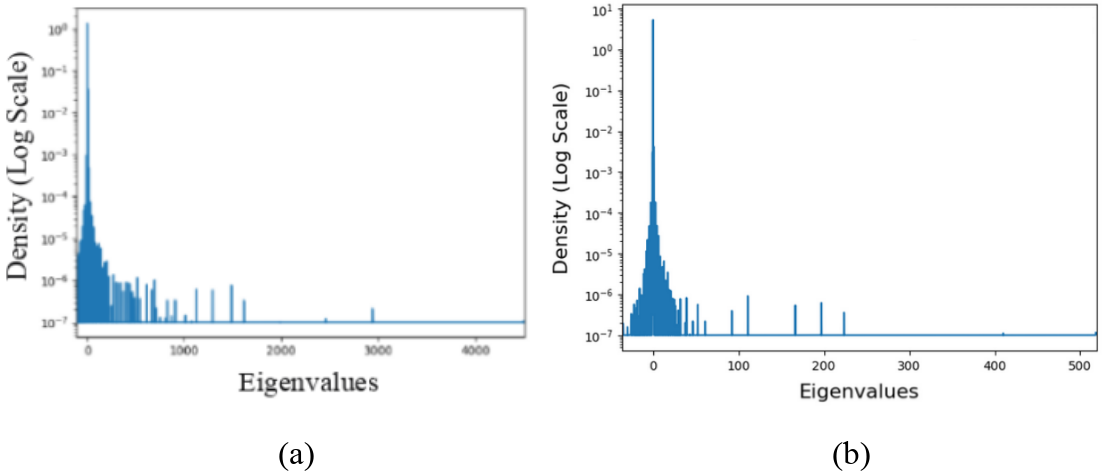}
    \caption{MP-HESDs of (a) ResNet18~\cite{RN} and (b) VIT small both trained on ImageNet.}\label{fig:2.1-1}
\end{figure}

Firstly, we verify that NNs typically have MP-HESD for a vast range of
NNs trained under different conditions. Secondly, we show that MN-HESD
is, in particular, a consequence of some external manipulation with the
gradients during NN training, which makes the proposed Hessian analysis
unreliable. Therefore, we conclude that the proposed HESD methodology~\cite{prev}
cannot be used to analyze NNs with MN-HESDs. Therefore, a
HESD type check becomes important, and we propose a HESD type criterion
along with a threshold value for HESD type determination. The proposed
criterion is combined with the previously proposed ones, forming a
unified generalization assessment methodology.

The rest of the paper is organized as follows: Section~\ref{typecrit} discusses HESD
types and formulates a MP-HESD criterion, Section~\ref{factors} studies the factors
that influence HESD type and discusses MN-HESD origins, Section~\ref{applic}
discusses the effects of HESD change during training on generalization
criteria and presents an updated generalization assessment algorithm,
and Section~\ref{conclusions} concludes the paper.

\section{HESD types and MP-HESD criterion}\label{typecrit}
\subsection{MP- and MN-HESDs}\label{MPvsMN}

It was observed in~\cite{prev} that HESDs of NNs are symmetrical upon
random initialization, with the negative section of HESD shrinking and
the positive one growing during training. This results in MP-HESDs of
trained NNs similar to those shown in Figure~\ref{fig:2.1-1}. Such HESDs have been
also reported in literature for CNNs~\cite{PyH} and Transformers~\cite{HGPT}, 
including Visual Transformers (VITs)~\cite{VT} and Generative
Pre-trained Transformers (GPTs)~\cite{GPT}, making them typical for a
large group of NNs.

\begin{figure} 
    \centering
    \includegraphics[scale=0.40]{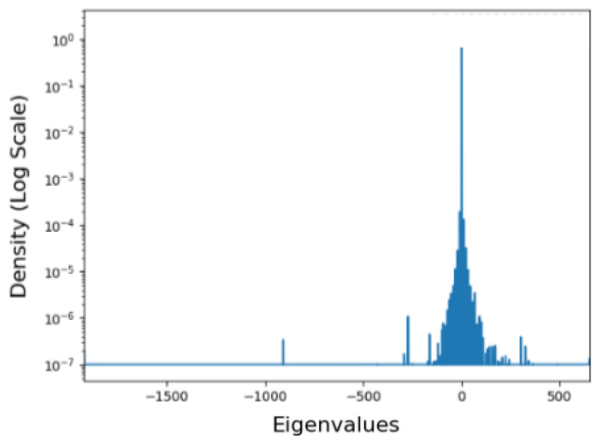}
    \caption{MN-HESD of VIT trained in~\cite{HTV}. Whereas large
    negative section typically suggests poor training, this VIT has 88\%
    training and 82\% generalization accuracies.}\label{fig:2.1-2}
\end{figure}

However, it was also reported in~\cite{prev} that NNs might have HESDs
which do not follow the above rules. Specifically, it was observed that
VIT trained in~\cite{HTV} has HESD shown in Figure~\ref{fig:2.1-2} which,
according to proposed criteria, indicates a poorly trained NN.
Nevertheless, this VIT had a rather high 88\% accuracy on ImageNet-1k~\cite{i1k}
and a remarkable 82\% generalization accuracy on ImageNet-v2~\cite{iv2}.

This observation gives rise to several important questions: what factors
lead to MN-HESDs formation, and do MN-HESDs always rules out the proposed
criteria deductions about good generalization on MP-HESDs only? To
answer these questions, we first need to take a look at the way specific
VITs were trained. There was a large group of VITs trained with a
specific goal of achieving high generalization accuracy. It involved
pre-training on augmented ImageNet-21k~\cite{i1k} with Adam~\cite{Adm}
followed by fine-tuning with SGD with momentum~\cite{GD} on ImageNet-1k.
In the following Section we will go through all training stages in~\cite{HTV}
to study their influence on HESD type: optimizer choice,
fine-tuning, data augmentation.

\subsection{MP-HESD criterion}\label{MPcrit}

To facilitate the HESD type assessment we propose the following
criterion

\begin{equation}
    C_{\text{t}} = \frac{\min(\lambda_{\text{neg}})}{\max(\lambda_{\text{pos}})}\
    \label{eq:2.2-1}
\end{equation}

where \(\lambda_{\text{neg}}\) and \(\lambda_{\text{pos}}\) are negative
and positive Hessian eigenvalues, respectively. Assuming the existence
of negative eigenvalues, this criterion is identical (up to a sign) to
generalization criterion \emph{r\textsubscript{e}} proposed in~\cite{prev}. 
Since random initialization HESDs are symmetrical and
negative section always decreases for MP- HESDs, one might assume that
the MP-HESD threshold value for \emph{C\textsubscript{t}} is -1.
However, it has been shown in~\cite{prev} that it can actually be higher for
some NNs (especially when BatchNorm and Dropout layers are present),
reaching as high as -0.73 for ResNet18 in evaluation mode. Therefore, it
is proposed to have a small safety margin and use -0.6 as the threshold
value, leading to the following condition for MP-HESD of trained NNs

\begin{equation}
    C_{\text{t}} > - 0.6\
    \label{eq:2.2-2}
\end{equation}

\section{Factors influencing HESD type}\label{factors}
\subsection{The influence of optimizer choice on HESD type}\label{optim}

\begin{figure} 
    \centering
    \includegraphics[scale=0.45]{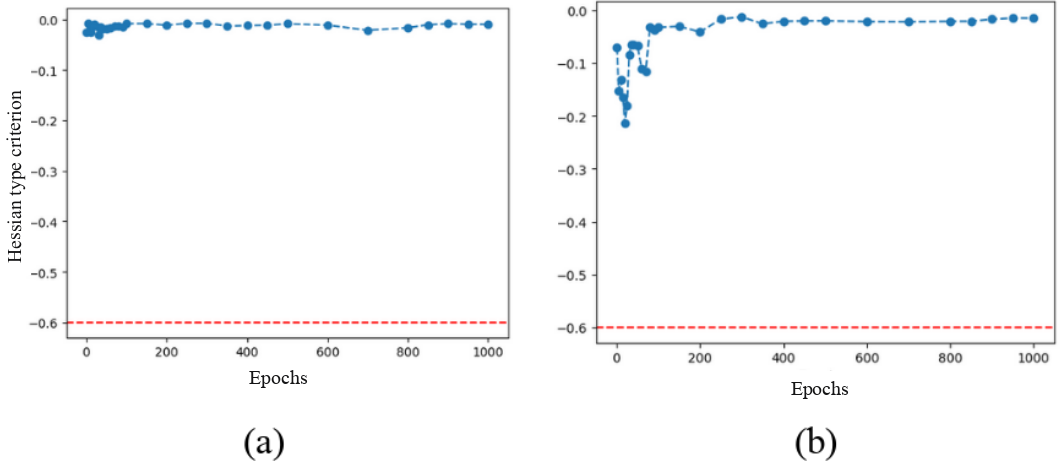}
    \caption{HESD criterion \emph{C\textsubscript{t}} values for (a)
    ResNet18 trained with AdamW and (b) VIT small trained with SGD with
    momentum on cinic. Points above the red line indicate belonging to
    MP-HESD type.}\label{fig:3.1-1}
\end{figure}

Several experiments involving training VIT small and ResNet20
on Cinic10~\cite{CIN} using SGD (with and without momentum) and Adam with
weight decay (AdamW~\cite{AW}) were conducted to study the influence of
optimizer choice on HESD type. Table~\ref{tab:3.1-1} summarizes the experiments
showing that the condition (\ref{eq:2.2-2}) is satisfied in all experiments,
indicating MP-HESDs. Figure~\ref{fig:3.1-1} shows that HESD type does not change
throughout training. The experimental procedure follows the one
described in Section 4 in~\cite{prev}. Hereafter we refer to Cinic10
without Cifar10 images as cinic~\cite{prev}, to Cifar10~\cite{CIF} as
cifar, to ImageNet-1k as i1k, and to ImageNet-v2 as iv2.

\begin{table}
    \caption{HESD type and accuracy for NN trained on different datasets
    with different optimizers.}\label{tab:3.1-1}
    \begin{tabular}{|l|l|l|l|l|l|l|}
        \hline
        Model & Training & Test & Optimizer & HESD type & Training & Generalization \\
        & dataset & dataset & & & accuracy, \% & accuracy, \% \\
        \hline
        ResNet20 & cinic & cifar & AdamW & MP & 99 & 55 \\ \hline
        ResNet20 & cinic & cifar & SGD mom & MP & 99 & 54 \\ \hline
        VIT small & cinic & cifar & AdamW & MP & 99 & 54 \\ \hline
        VIT small & cinic & cifar & SGD & MP & 99 & 43 \\ \hline
        VIT small & cinic & cifar & SGD mom & MP & 99 & 50 \\ \hline
    \end{tabular}
\end{table}

\subsection{Training and fine-tuning}\label{traintune}

Training models on ImageNet-1k with AdamW and fine-tuning them on cinic
with SGD with momentum (following the procedure in~\cite{HTV}) was
performed to investigate whether fine-tuning from pre-trained state
might result in MN-HESDs. Since input size and the number of classes are
different in the studied datasets, both input and classifier layers were
altered in NNs after pre-training. Fine-tuning was conducted in two
regimes: fine-tuning all weights with low learning rate, and freezing
all pre-trained weights except those belonging to modified layers.

\begin{figure} 
    \centering
    \includegraphics[scale=0.5]{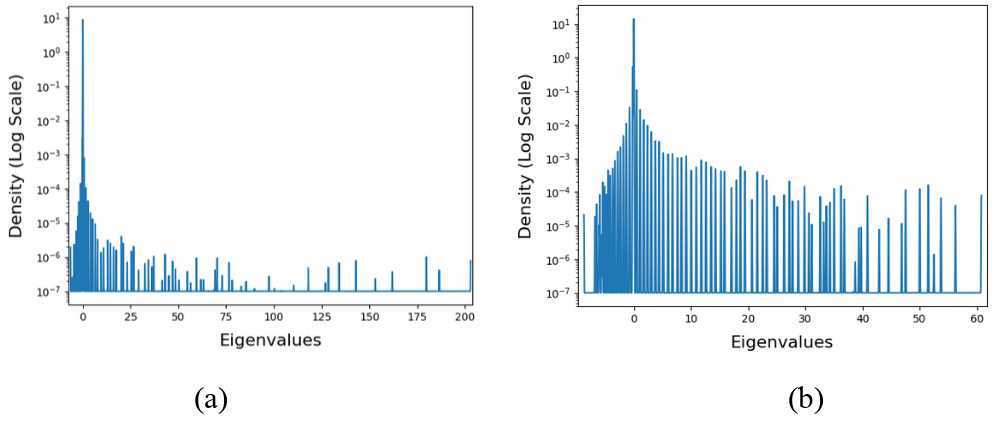}
    \caption{HESD of VIT after fine-tuning (a) all and (b) only input
    and classifier layer weights.}\label{fig:3.2-1}
\end{figure}

The results summarized in Table~\ref{tab:3.2-1} show MP-HESDs for all experiments.
Figure~\ref{fig:3.2-1} shows that HESDs are different when pre-trained weights are
frozen compared to full fine-tuning. In our experiments the latter
results in higher generalization accuracy.

\begin{table}
    \caption{HESD type and accuracy for VIT trained on i1k, fine-tuned
    on cinic and tested on cifar.}\label{tab:3.2-1}
    \begin{tabular}{|l|l|l|l|l|l|l|l|l|}
        \hline
        Model & Training & Tuning & Test & Optimizer & Frozen & HESD & Train & Gen. \\
        & dataset & dataset & dataset & & weights & type & acc., \% & acc., \% \\
        \hline
        VIT & i1k & cinic & cifar & SGD & no & MP & 99 & 35 \\ \hline
        VIT & i1k & cinic & cifar & SGD & yes & MP & 84 & 30 \\ \hline
        VIT & i1k & cinic & cifar & SGD mom & no & MP & 99 & 40 \\ \hline
        VIT & i1k & cinic & cifar & SGD mom & yes & MP & 91 & 31 \\ \hline
    \end{tabular}
\end{table}

\subsection{Data augmentation}\label{augmentation}

Following the methodology in~\cite{HTV}, RandAugment~\cite{aug} and
specifically aug\_light1 was used for data augmentation in our
experiments. Table~\ref{tab:3.3-1} shows MP-HESDs for all experiments. It
also shows that generalization accuracy increases significantly when
data augmentation is used, verifying the results reported in~\cite{HTV}.

\begin{table}
    \caption{HESD type and accuracy for VIT trained with data
    augmentation.}\label{tab:3.3-1}
    \begin{tabular}{|l|l|l|l|l|l|l|}
        \hline
        Model & Training & Test & Optimizer & HESD type & Training & Generalization \\
        & dataset & dataset & & & accuracy, \% & accuracy, \% \\
        \hline
        VIT small & cinic & cifar & AdamW & MP & 99 & 75 \\ \hline
        VIT small & cinic & cifar & SGD & MP & 95 & 60 \\ \hline
        VIT small & cinic & cifar & SGD mom & MP & 97 & 73 \\ \hline
        VIT Small & i1k & iv2 & AdamW & MP & 89 & 46 \\ \hline
    \end{tabular}
\end{table}

\subsection{MN-HESD of VIT trained with AdaHessian}\label{adahessian}

The inability to obtain MN-HESD in previous experiments has inspired us
to conduct experiments with an optimizer which is directly linked to
Hessian analysis -- AdaHessian~\cite{AdaH}. It uses Hessian diagonal
approximation for gradient rescaling and rotation in weight update. This
is similar to using Hessian axes to choose the optimization direction in
weight space when traversing the loss landscape of NNs~\cite{PyH,Hd}.

\begin{figure} 
    \centering
    \includegraphics[scale=0.35]{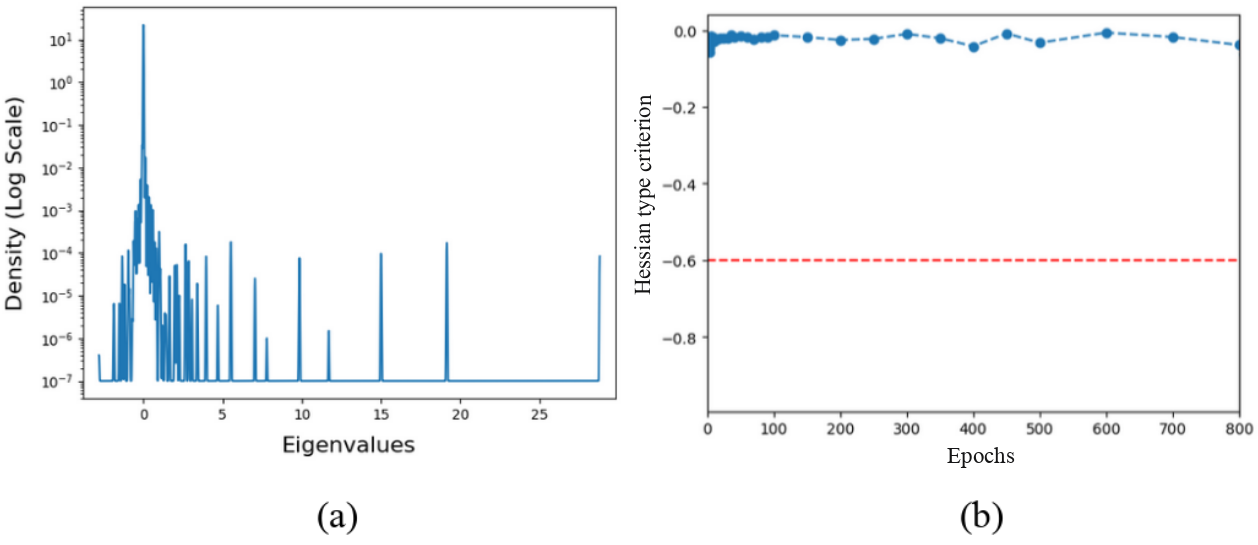}
    \caption{(a) MP-HESD of ResNet20 trained on cinic with AdaHessian
    and (b) training \emph{C\textsubscript{t}}.}\label{fig:3.4-1}
\end{figure}

\begin{figure} 
    \centering
    \includegraphics[scale=0.40]{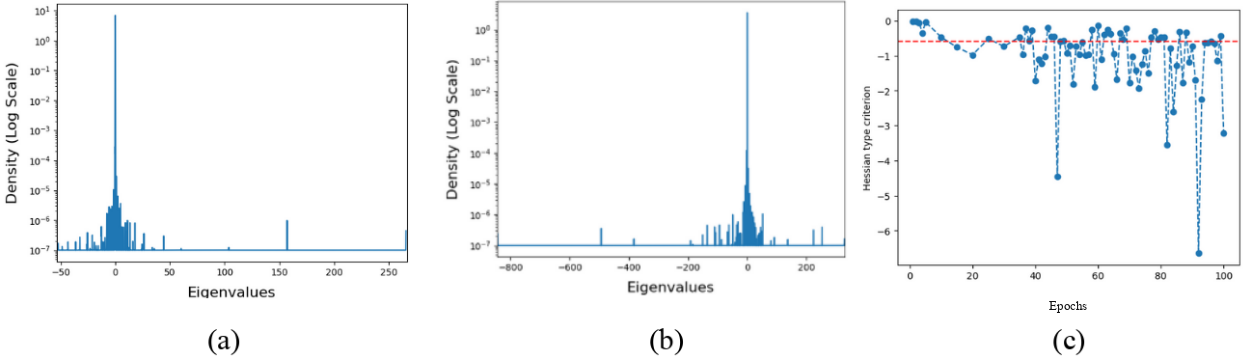}
    \caption{HESD types of VIT trained on cinic with AdaHessian: (a)
    MP-HESD for epoch 86, (b) MN-HESD for epoch 87, and (c) changes in
    \emph{C\textsubscript{t}} with epochs. Every point below the red
    threshold line in (c) indicates MN-HESD.}\label{fig:3.4-2}
\end{figure}

Figure~\ref{fig:3.4-1} shows HESD and \emph{C\textsubscript{t}} for ResNet20
trained on cinic, indicating MP-HESD throughout training. However, this
is not the case for VIT trained on cinic. Figure~\ref{fig:3.4-2} (a) shows MP-HESD
for some epochs. However, Figure~\ref{fig:3.4-2} (b) shows that MN-HESD can occur
for a neighboring epoch. Figure~\ref{fig:3.4-2} (c) further illustrates this by
showing how \emph{C\textsubscript{t}} and HESD type change between
epochs.

Table~\ref{tab:3.4-1} also shows that generalization accuracy of VIT trained with
AdaHessian is lower than that obtained with AdamW. This means that
MN-HESD is not an indicator of NN generalization capabilities, answering
one of the main questions posed in Section~\ref{introduction}.

\begin{table}
    \caption{HESD type and classification accuracies of VITs trained
    with different optimizers.}\label{tab:3.4-1}
    \begin{tabular}{|l|l|l|l|l|l|l|}
        \hline
        Model & Training & Test & Optimizer & HESD type & Training & Generalization \\
        & dataset & dataset & & & accuracy, \% & accuracy, \% \\
        VIT small & cinic & cifar & AdamW & MP & 99 & 54 \\ \hline
        VIT small & cinic & cifar & SGD & MP & 99 & 43 \\ \hline
        VIT small & cinic & cifar & SGD mom & MP & 99 & 50 \\ \hline
        VIT small & cinic & cifar & AdaHessian & MN & 99 & 49 \\ \hline
    \end{tabular}
\end{table}

However, it remains unclear what influences the type of HESD. The hint
is in the difference between ResNet20 and VIT training with AdaHessian,
which includes spatial averaging in Hessian diagonal approximation
\emph{D} calculation~\cite{AdaH}. This is done to reduce noise
sensitivity of the second order optimization algorithms. For CNNs this
results in averaging on a scale of convolutional filters, which are
commonly 3x3, 5x5, or 7x7. However, for VITs this leads to averaging
over the hidden dimension, which in case of VIT small is 384. This leads
to individual gradients, and their values being obtained after
AdaHessian averaging, being significantly different for attention
layers, which are major contributors to the total HESD~\cite{HGPT}. Since
HESD evaluation algorithm does not account for optimizer operations, the
gradients it uses give a perspective very different from the expected
one. It should be stressed that this effect has no tangible influence on
normal operating regimes of NNs with MN-HESDs, since gradients are not
used for inference.

Therefore, MN-HESDs are a consequence of external gradient manipulation.
While it cannot be verified exactly, we conclude that in case of MN-HESD
VIT from~\cite{HTV} gradient clipping at global norm 1 results in
HESD distortion and MN-HESD for some epochs. Similar to the previously
discussed case of AdaHessian, it leads to gradients used by the
optimizer and the ones used in Hessian calculations being differents.
This observation can be generalized further, meaning that Hessian
analysis methodology becomes less reliable when any operation that
significantly affects gradients is used during training.

\begin{figure} 
    \centering
    \includegraphics[scale=0.38]{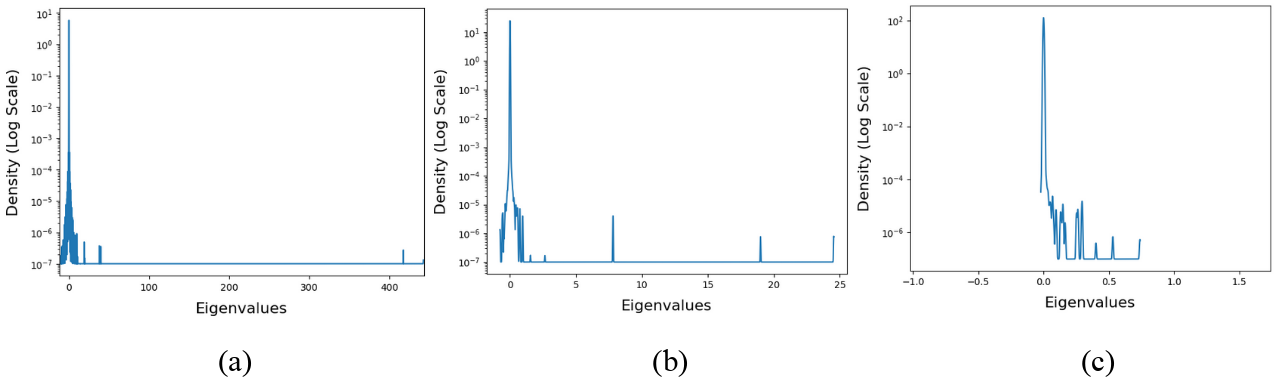}
    \caption{HESDs of VIT trained on cinic with AdamW at (a) 100th, (b)
    250th, and (c) 450th epoch.}\label{fig:4.1-1}
\end{figure}

\section{Applicability of the HESD generalization criteria}\label{applic}
\subsection{Excessive training and quasi-singular HESD}\label{qs-hesd}

\begin{figure} 
    \centering
    \includegraphics[scale=0.4]{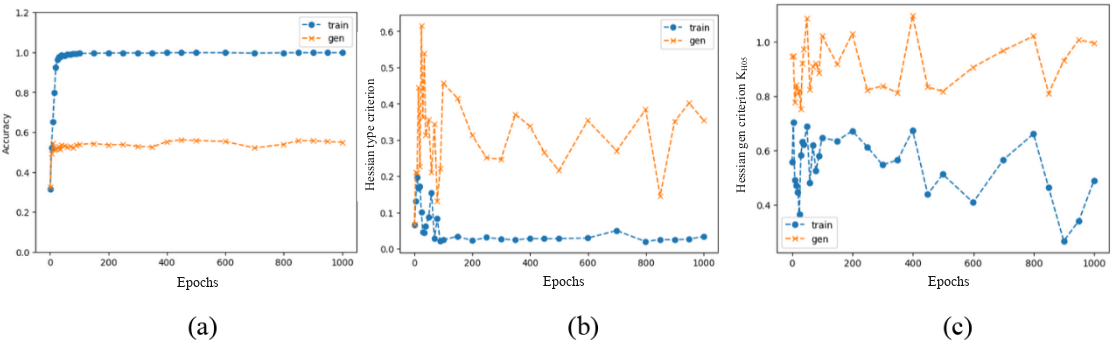}
    \caption{Changes in Hessian generalization criteria~\cite{prev} with
    epochs for VIT trained on cinic with AdamW, (a)
    \emph{r\textsubscript{e}} and (b) \emph{K\textsubscript{H05}}.}\label{fig:4.1-2}
\end{figure}

Whereas previous results show that MP-HESD are typical for the majority
of studied cases, there is an additional peculiarity that occurs when NN
is being trained after final accuracy has already been reached. This
leads to eigenvalues decreasing with epochs, an effect that has
previously been theorized in~\cite{4H}. The results shown in Figure~\ref{fig:4.1-1}
indicate that eigenvalues indeed become smaller and HESD width decreases
during excessive training. This results in quasi-singular (QS) HESD with
all of its eigenvalues tending to zero (see Figure~\ref{fig:4.1-1} (c)). However,
training and generalization accuracies of cinic-trained VIT shown in
Figure~\ref{fig:4.1-2} (a) remaining practically unchanged.

However, HESD generalization criteria \emph{r\textsubscript{e}} and
\emph{K\textsubscript{H05}}~\cite{prev} are not actually affected by
these HESD changes. Figures~\ref{fig:4.1-2} (b) and (c) show that while the
criteria values do change, their changes for train and generalization
datasets are proportional, meaning that their difference always
indicates poor generalization (which is indeed the case, as shown in
Figure~\ref{fig:4.1-2} (a)). Nevertheless, the changes in HESD shown in Figure~\ref{fig:4.1-1}
could also motivate future research of criteria that are not
affected or are less susceptible to the changes in HESDs with epochs.

\subsection{Comparing maximum eigenvalue and \emph{C\textsubscript{t}}
as generalization criteria}\label{gencrit}

\begin{figure} 
    \centering
    \includegraphics[scale=0.35]{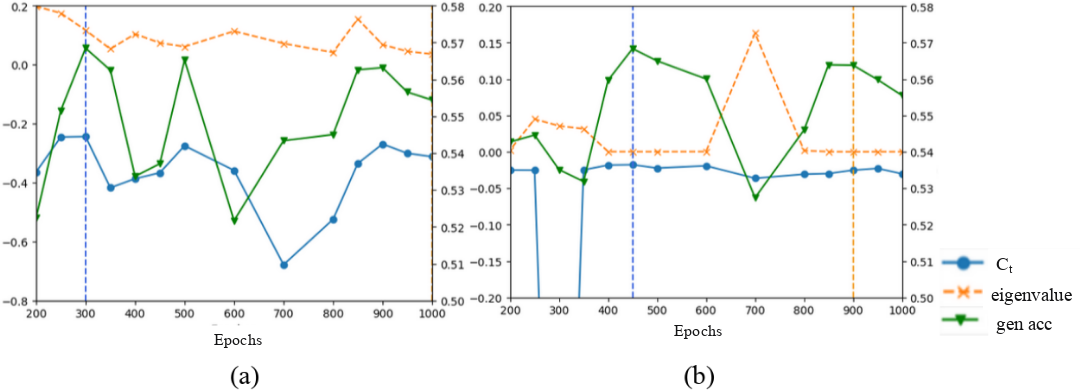}
    \caption{Maximum eigenvalue, \emph{C\textsubscript{t}}, and
    generalization accuracy of (a) ResNet20 and (b) VIT trained on cinic and
    generalized to cifar.}\label{fig:4.2-1}
\end{figure}

Maximum Hessian eigenvalue, and sometimes Hessian trace, are
conventionally believed to be good indicators of loss landscape flatness
and generalization capability of NNs~\cite{LLO, 4H}. That is, the smaller
the maximum eigenvalue the more flat the loss landscape becomes, which
has been shown to lead to better generalization~\cite{LLO}. However, the
previous Section shows that eigenvalues can decrease drastically at
nearly constant generalization accuracy simply due to excessive
training. That is, one could hypothetically obtain any eigenvalues or
trace of Hessian matrix without affecting NN's generalization
capability, as Figures~\ref{fig:4.1-1} and~\ref{fig:4.1-2} show.

On the contrary, the proposed criterion \emph{C\textsubscript{t}} does
not become degenerate for QS-HESDs, and it is more stable between
epochs. It also can be used to choose weights that provide better
generalization: among weights with nearly the same training accuracy,
the ones with the highest \emph{C\textsubscript{t}} should be chosen.
Geometrically this means choosing a model which loss landscape has the
least \emph{ratio} of negative and positive curvatures.

Figure~\ref{fig:4.2-1} compares the use of lowest maximum eigenvalue and maximum
\emph{C\textsubscript{t}} criteria for weight choice for ResNet20 and
VIT trained with AdamW on cinic and tested on cifar. Figure~\ref{fig:4.2-1} (a)
shows that eigenvalue method predicts best result for
1000\textsuperscript{th} epoch which provides 55.4\% generalization
accuracy. However, \emph{C\textsubscript{t}} method predicts best
results for 300\textsuperscript{th} epoch with 56.9\% generalization
accuracy. Here the proposed method allows to choose weights with 1.5\%
better generalization accuracy at an earlier epoch.

Similarly, Figure~\ref{fig:4.2-1} (b) shows 56.4\% generalization accuracy
obtained using eigenvalue method at 900\textsuperscript{th} epoch and
56.8\% generalization accuracy for 450\textsuperscript{th} epoch chosen
using the \emph{C\textsubscript{t}} method. Whereas these results are
close, \emph{C\textsubscript{t}} method is still slightly more accurate
when finding weights with better generalization at earlier epochs.

\begin{figure} 
    \centering
    \includegraphics[scale=0.35]{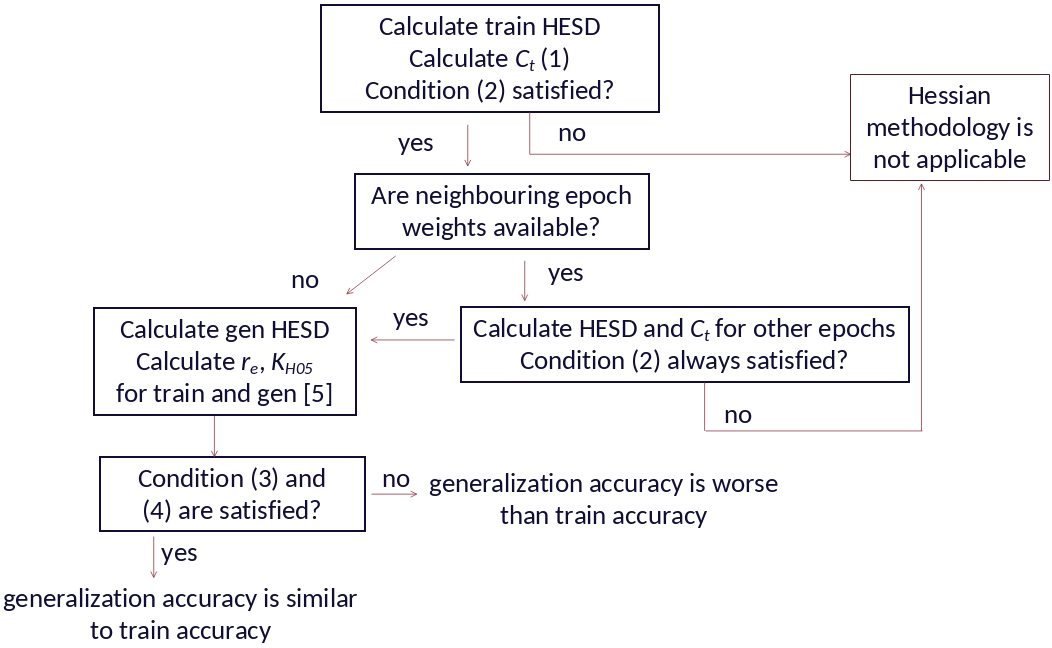}
    \caption{Hessian generalization criteria application algorithm.}\label{fig:4.3-1}
\end{figure}

\subsection{Updated HESD generalization assessment methodology}\label{method-upd}

It was previously discussed in~\cite{prev} how changes in HESD
generalization criteria can be used to assess generalization
capabilities of NNs. In this Section we also provide numerical
conditions that should be satisfied to indicate good generalization

\begin{equation}
    \mathrm{\Delta}r_{e} = \frac{r_{e.\text{gen}}}{r_{e.\text{train}}} < 1.5\
    \label{eq:4.2-1}
\end{equation}

\begin{equation}
    \mathrm{\Delta}K_{H05} = \frac{K_{H05.\text{gen}}}{K_{H05.\text{train}}} < 1.2\
    \label{eq:4.2-2}
\end{equation}

Since HESD generalization criteria rely heavily on the fact that the
ratio between negative and positive HESD sections can be used as an
indicator of NN training and generalization quality, they can only be
applied to MP-HESD NNs. Therefore, HESD type check (\ref{eq:2.2-2}) is needed to
determine whether criteria (\ref{eq:4.2-1}) and (\ref{eq:4.2-2}) can be applied to analyze
NN generalization. Figure~\ref{fig:4.3-1} summarizes the updated algorithm for
Hessian-based generalization capability assessment.

\section{Conclusions}\label{conclusions}

This paper has shown that determining HESD type is essential for
assessing the applicability of HESD analysis to studying generalization
capabilities of NNs. It has been shown that MP-HESDs are typical in
training and fine-tuning and are not affected by the choice of NN
classifier model, optimizer, dataset, preprocessing, and augmentation
methods. MN-HESDs occur when external gradient manipulation takes place,
such as averaging, normalization, or rescaling of large groups of
gradients. Since the extent of such manipulations is hard to predict,
the proposed Hessian methodology becomes unreliable and its use is not
recommended in these cases. However, HESD generalization criteria can
provide valuable insight into the generalization capabilities of MP-HESD
NNs. In this paper HESD type criterion is proposed which, when combined
with previously proposed generalization assessment criteria, completes
the NN generalization assessment method for MN-HESD NNs.

\section*{Acknowledgement}\label{acknowledgement}

The author would like to thank his colleagues Dr Anton Raskovalov, Dr
Igor Netay, and Ilya Androsov for fruitful discussions, and Vasily
Dolmatov for discussions and project supervision.

\bibliographystyle{IEEEtran}
\bibliography{IEEEabrv,ms}

\end{document}